\def\year{2019}\relax
\documentclass[letterpaper]{article} 
\usepackage{aaai}  
\usepackage{times}  
\usepackage{helvet}  
\usepackage{courier}  
\usepackage{url}  
\usepackage{graphicx}  

\usepackage{latexsym}
\usepackage{graphics}  
\usepackage{subcaption}
\usepackage{caption}
\usepackage{color}
\usepackage{comment}
\usepackage{booktabs}
\usepackage{textcomp}
\usepackage{amsmath}
\usepackage{verbatim}
\usepackage{amsmath}

\usepackage{marginnote}

\usepackage{algorithm}
\usepackage{algpseudocode}
\usepackage{amsmath}

\begin{document}
%
\title{ Incorporating Structured Commonsense Knowledge in Story Completion }

\author{
Jiaao Chen \\ Computer Science Department \\ Zhejiang University \\ Hangzhou, China \And
Jianshu Chen \\ Tencent AI Lab \\   Bellevue, WA, USA \And
Zhou Yu \\ Computer Science Department \\  University of California, Davis \\ Davis, CA, USA
}

\maketitle
\begin{abstract}
The ability to select an appropriate story ending is the first step towards perfect narrative comprehension. Story ending prediction requires not only the explicit clues within the context, but also the implicit knowledge (such as commonsense) to construct a reasonable and consistent story. However, most previous approaches do not explicitly use background commonsense knowledge. We present a neural story ending selection model that integrates three types of information: narrative sequence, sentiment evolution and commonsense knowledge. Experiments show that our model outperforms state-of-the-art approaches on a public dataset, ROCStory Cloze Task \cite{W17-0906}, and the performance gain from adding the additional commonsense knowledge is significant. 
\end{abstract}

\section{Introduction}
Narrative is a fundamental form of representation in human language and culture. Stories connect individuals and deliver experience, emotions and  knowledge. Narrative comprehension has attracted long-standing interests in natural language processing (NLP) \cite{history}, and is widely applicable to areas such as content creation. Enabling machines to understand narrative is also an important first step towards real intelligence. Previous studies on narrative comprehension include character roles identification \cite{valls2015ijcai},  narratives schema construction \cite{Chambers:2009:ULN:1690219.1690231}, and plot pattern identification \cite{Jockers:2013:MDM:2530314}. However, their main focus is on analyzing the stories themselves. In contrast, we concentrate on training machines to predict the end of the stories.  Story completion tasks rely not only on the logic of the story itself, but also requires implicit \emph{commonsense knowledge} outside the story. To understand stories, human can use the information from both the story itself and other implicit sources such as commonsense knowledge and normative social behaviors \cite{Mueller:2003:SUT:1119239.1119246}. In this paper, we propose to imitate such behaviors to incorporate structured commonsense knowledge to aid the story ending prediction.

Recently, \cite{W17-0906} introduced a ROCStories dataset as a benchmark for evaluating models' ability to understand the narrative structures of a story, where the model is asked to select the correct ending from two candidates for a given story. To solve this task, both traditional machine learning approaches \cite{schwartz-EtAl:2017:LSDSem} and neural network models \cite{P17-2097} have been used. Some works also exploit information such as sentiment and topic words \cite{D17-1168} and event frames \cite{C18-1149}. Recently, there has been work \cite{openai} that leverages large unlabeled corpus, like the BooksCorpus \cite{journals/corr/ZhuKZSUTF15} dataset, to improve the performance. However, none of them explicitly uses structured commonsense knowledge, which humans would naturally incorporate to improve model performance.

\begin{figure}[t]
    \centering
    \begin{subfigure}[t]{0.45\textwidth}
        \centering
        \includegraphics[width=0.8\textwidth]{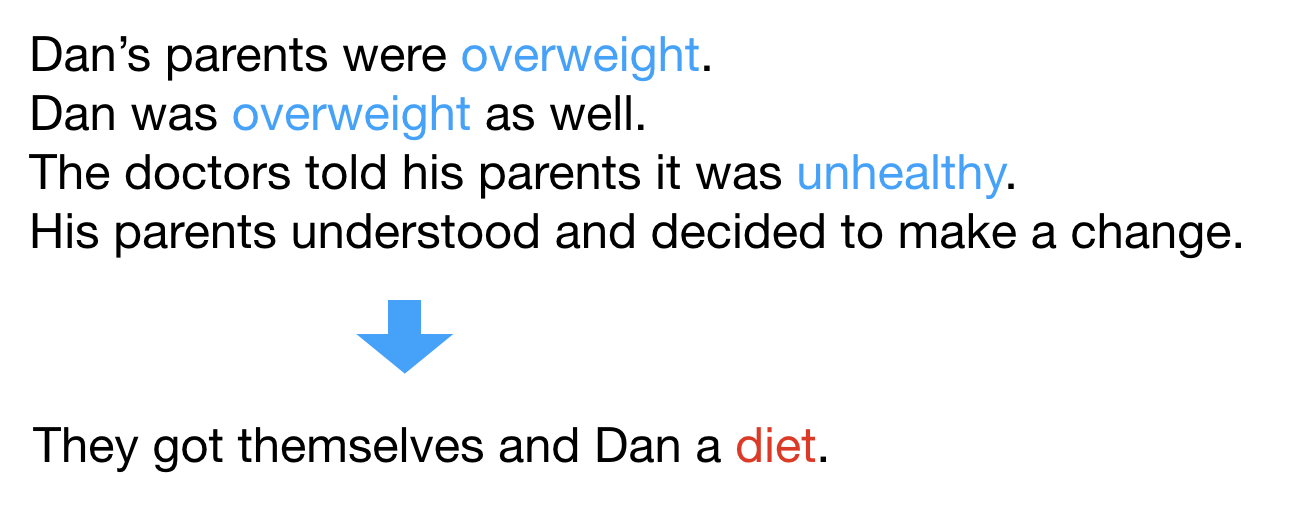}
        \caption{An example story }\label{example_story}
    \end{subfigure}
    ~
    \begin{subfigure}[t]{0.45\textwidth}
        \centering
        \includegraphics[width=1.0\textwidth]{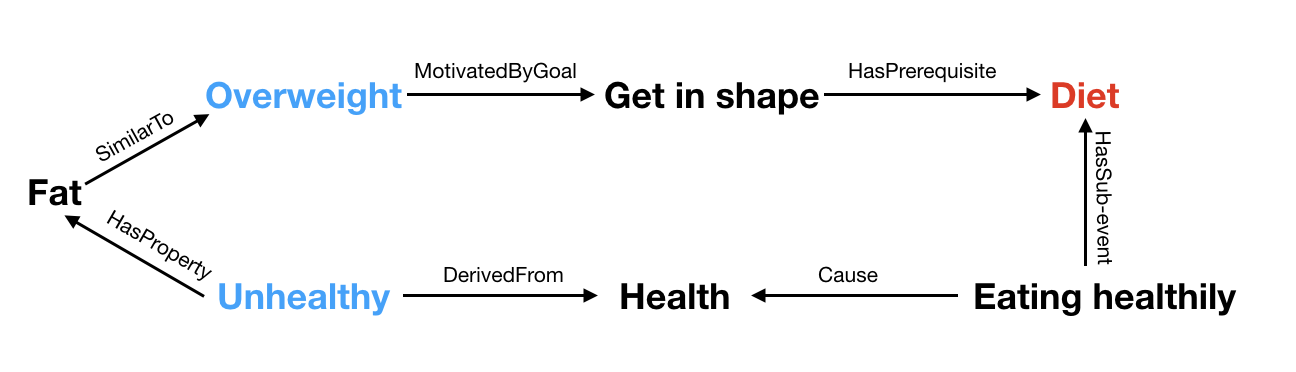}
        \caption{Clues in ConceptNet}\label{KG}
    \end{subfigure}
    \caption{(a) shows an example story from ROCStories dataset, words in colors are key-words. (b) shows the key-words and their relations in ConceptNet Knowledge Graph}\label{Example}
\end{figure}

\begin{figure*}[tp]
\centering
\includegraphics[width=0.7\textwidth]{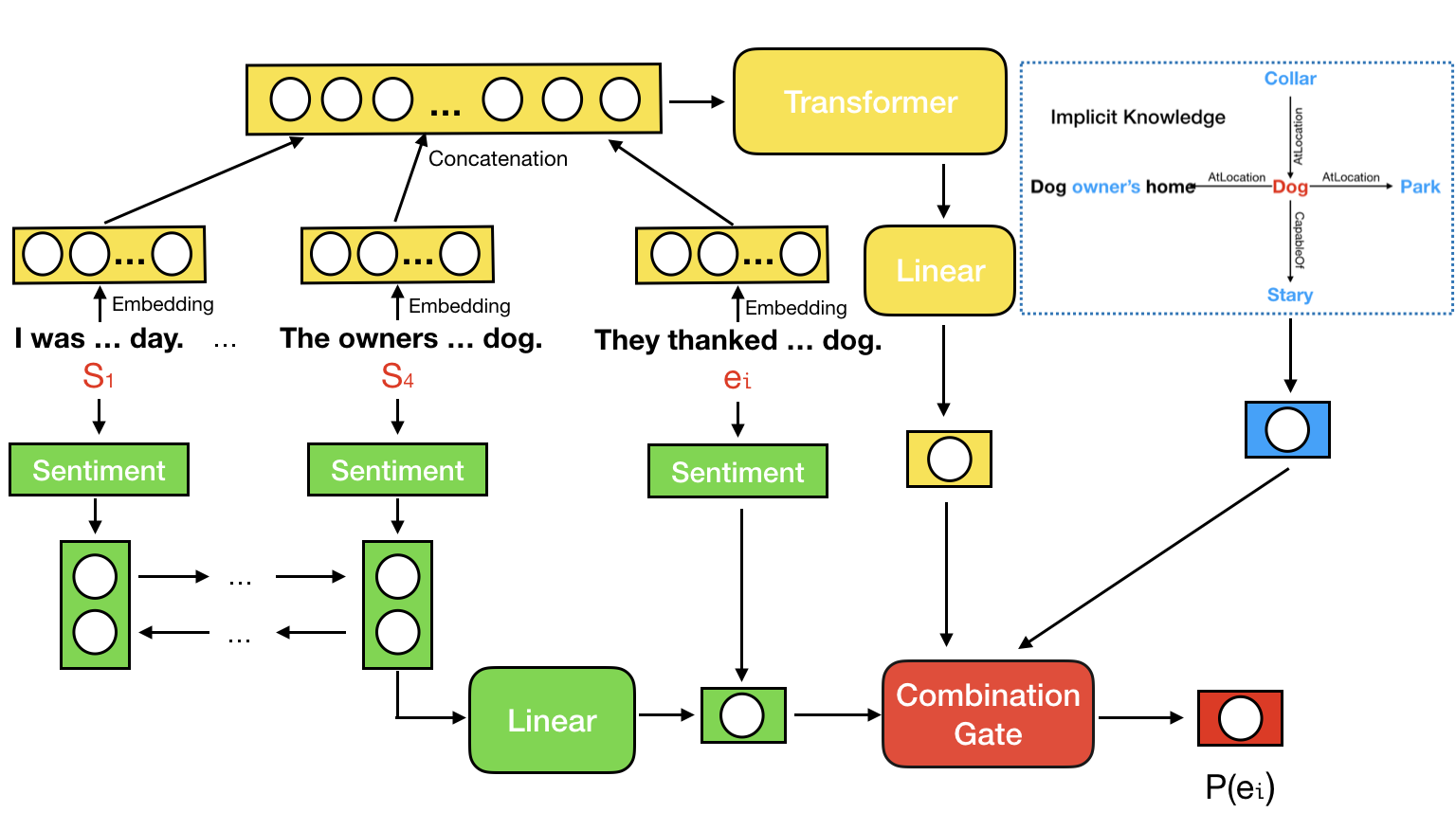}
\caption{Our proposed model architecture. The inputs: $S_1$ through $S_4$ denote the story body, and $e_i$ ($i=1,2$) denotes two candidate endings. The bottom-left  component encodes sentiment evolution information (green), the top-left component models the narrative sequence (yellow), and the top-right component  integrates commonsense knowledge (blue). The combination gate in the bottom-right integrates all three types of information and outputs the probability on which ending is correct.}\label{Models}
\end{figure*}

Figure \ref{Example}(a) shows a typical example in ROCStories dataset: a story about Dan and his parents. The blue words are key-words in the body of the story, and the red word is the key-word in the correct story ending. Figure \ref{Example}(b) shows the (implicit) relations among these key-words, which are obtained as a subgraph from ConceptNet \cite{speer2017conceptnet}, a commonsense knowledge base. By incorporating such structured external commonsense knowledge, we are able to discover strong associations between these keywords and correctly predict the story ending. Note that these associations are not available from the story itself.

To solve the story completion task, we propose a neural network model that integrates three types of information: (i) narrative sequence, (ii) sentiment evolution, and (iii) commonsense knowledge. The clues in narrative chain  are captured by a transformer decoder, constructed from a pretrained language model. The sentiment prediction is obtained by using a LSTM model. Additionally, the commonsense knowledge is extracted from an existing structured knowledge base, ConceptNet. In particular, we use a combination gate to integrate all the information and train the model in an end-to-end manner. Experiments demonstrate the improved performance of our model on the task. 

\section{Related Work}
Our work on story completion is closely related to several research areas such as reading comprehension, sentiment analysis and commonsense knowledge integration, which will be briefly reviewed as below.

\textbf{Reading Comprehension} is the ability to process text, understand its meaning, and to integrate it with what the readers already know. It has been an important field in NLP for a long time. The SQuAD dataset \cite{DBLP:conf/acl/RajpurkarJL18} presents a task to locate the correct answer to a question in a context document and recognizes unanswerable questions. The RACE dataset \cite{lai-EtAl:2017:EMNLP2017}, which is constructed from Chinese Students English Examination, introduces another task that requires not only retrieval but also reasoning. Usually they are solved by match-based model like QANET \cite{DBLP:journals/corr/abs-1804-09541}, hierarchical attention model like HAF \cite{AAAI1816331}, and dynamic fusion based model like DFN \cite{Xu2018Dynamic}. Also there exists more relevant research on story comprehension such as event understanding of narrative plots \cite{Chambers:2009:ULN:1690219.1690231}, character personas \cite{valls2015ijcai} and inter-character relationships \cite{Iyyer2016Feuding}.

\textbf{Sentiment Analysis} aims to determine the attitude of a speaker (or a writer) with respect to some topic, the overall contextual polarity, or emotional reaction to a document, interaction or event. There have been rich studies on this field, such as learning word vectors for sentiment analysis \cite{Maas:2011:LWV:2002472.2002491} and recognizing contextual polarity in a phrase-level \cite{Wilson:2005:RCP:1220575.1220619}. Recently, researchers studied large-scale sentiment analysis across news and blogs \cite{Godbole2007Large}, and also studied opinion mining on twitter \cite{Patodkar2010Twitter}. Additionally, there have been studies focused on joint learning for better performance, such as detecting sentiment and topic simultaneously from text \cite{Lin:2009:JSM:1645953.1646003}.

\textbf{Commonsense Knowledge Integration} If machines receive information from a commonsense knowledge base, they become more powerful for many tasks like reasoning \cite{Bagherinezhad2016Are}, dialogue generation \cite{DBLP:conf/acl/LiuFCRYL18} and cloze style reading comprehension \cite{Mihaylov2018Knowledgeable}. Related works include \cite{Bagherinezhad2016Are}, which builds a knowledge graph and uses it to deduce the size of objects \cite{Bagherinezhad2016Are}, in addiiton to \cite{Zhu2017Flexible}, in which a music knowledge graph is built for a single round dialogue system. There are several ways to incorporate external knowledge base (e.g., ConceptNet). For example, \cite{DBLP:journals/corr/SpeerL17} uses a knowledge based word embedding, \cite{Young2017Augmenting} employs tri-LSTMs to encode the knowledge triple, and \cite{Zhou2018Commonsense} and \cite{Mihaylov2018Knowledgeable} apply graph attention embedding to encode sub-graphs from a knowledge base. However, their work does not involve narrative completion.

\textbf{Story Completion}
Traditional machine learning methods have been used to solve ROCStory Cloze Task  such as \cite{schwartz-EtAl:2017:LSDSem}. To improve the performance, features like topic words and sentiment score are also extracted and incorporated \cite{D17-1168}. Neural network models have also been applied to this task (e.g., \cite{learning} and \cite{P17-2097}), which use LSTM to encode different parts of the story and calculate their similarities. In addition, \cite{C18-1149} introduces event frame to their model and leverages five different embeddings. Finally, \cite{openai} develops a transformer model and achieves state-of-the-art performance on ROCStories, where the transformer was pretrained on BooksCorpus (a large unlabeled corpus) and finetuned on ROCStories.

\section{Proposed Model}
For a given story $S = \{s_1, s_2, ..., s_L\}$ consisting of a sequence of $L$ sentences, our task is to select the correct ending out of two candidates, $e_1$ and $e_2$, so that the completed story is reasonable and consistent. On the face of it, the problem can be understood as a standard binary classification problem. However, learning binary classifier with standard NLP techniques on the explicit information in the story is not sufficient. This is because correctly predicting the story ending usually requires reasoning with implicit commonsense knowledge. Therefore, we develop a neural network model to predict the story ending by integrating three sources of information: narrative sequence, sentiment evolution and structured commonsense knowledge (see Figure \ref{Models}). Note that the first two types of information are explicit in the story while the third type is implicit and has to be imported from external source such as a knowledge base. In this section, we will explain how we exploit these three information sources and integrate them to make the final prediction.

\subsection{Narrative Sequence}
\begin{figure}[t]
\centering
\includegraphics[width=6cm,height=3cm]{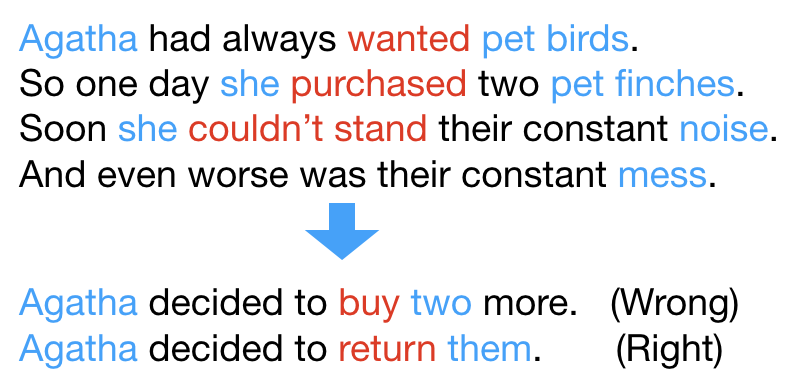}
\caption{An example story in the ROCStories dataset}\label{Example1}
\end{figure}

To describe a consistent story, plots should be planned in a logically reasonable sequence; that is there should be a narrative chain between different characters in the story. This is illustrated in the example in Figure \ref{Example1}, where words in red are events and words in blue are characters. The story chain, ``\emph{Agatha wanted pet birds} $\rightarrow$ \emph{Agatha purchased pet finches} $\rightarrow$ \emph{Agatha couldn't stand noise} $\rightarrow$ \emph{mess was worse} $\rightarrow$ \emph{Agatha return pet birds}", describes a more coherent and reasonable story than `` \emph{Agatha wanted pet birds} $\rightarrow$ \emph{Agatha purchased pet finches} $\rightarrow$ \emph{Agatha couldn't stand noise} $\rightarrow$ \emph{mess was worse} $\rightarrow$ \emph{Agatha buy two more}". When Agatha could not stand the noise, it is more likely for her to give these birds away rather than buy more. Therefore, developing a better semantic representation for narrative chains is important for us to predict the right endings. 

Inspired by the recent research from OpenAI \cite{openai} on forming semantic representations of narrative sequences, we first pre-train a high-capacity language model on a large unlabeled corpus of text to learn the general information hidden in the context, and then fine-tune the model on this story completion task. 

Given a large corpus of tokens $C = \{c_1, c_2, ... , c_n\}$, we can pre-train a language model to maximize the likelihood :
\begin{equation}
    L_{lm}(C) = \sum_{i} \log P_l(c_i|c_{i-k}, ..., c_{i-1}; \theta)
\end{equation}
where $k$ is the window size, and the conditional probability $P_l$ is modeled using a neural network with parameters $\theta$.

Similar to \cite{openai}, we use a multi-layer transformer decoder with multi-headed self-attention for the language model:
\begin{align}
    h_0 &= C W_e+W_p  \\
    h_l &= transformer(h_{l-1}),  l \in[1,M] \\
    P(c) &= softmax(h_M W_e^T) 
\end{align}
where $C = \{c_1, c_2, ... , c_n\}$ are tokens in corpus, $W_e$ is the token embedding matrix, $W_p$ is the position embedding matrix and $M$ is the number of transformer blocks. 

We use the pre-trained parameters released by OpenAI \footnote{\url{https://github.com/openai/finetune-transformer-lm}} as the initialization for the transformer decoder. We adapt these parameters to our classification task. For each candidate story $(s_1, s_2, s_3, s_4, e_i)$ (i.e., the story body followed by one candidate ending), we serialize it into a sequence of tokens $X = \{x_1, ... , x_k\}$, where $k$ is the number of tokens. Then the fine-tuned transformer takes $X$ as its input and outputs the probability of $e_i$ being the correct ending:
\begin{equation}
    \label{Equ:P_N}
   P_N(y|s_1, ..., s_4, e_i) = softmax(W_M h_M^k + b_M)
\end{equation}
where $y \in \{0,1\}$ is the label indicating whether $e_i$ is the correct ending, $h_M^k$ denotes the hidden representation at the $M$-th layer of the transformer associated with the $k$-th token, and $W_M$ and $b_M$ are parameters in the linear output layer.

\subsection{Sentiment Evolution}
Besides narrative sequence, getting a good sentiment prediction model is also important for choosing the correct endings. Note that stories are different from other objective texts (e.g., news), as they have emotions within the context. Usually there is a sentiment evolution when a storyline is being revealed \cite{sentiment} .

First, we pre-train a sentiment prediction model using the training set of the ROCStories, which does not have alternative endings (i.e., no negative samples). Given a five-sentence story $S = \{s_1, s_2, s_3, s_4, s_5\}$, we take the first four sentences as the body $B$ and the last sentence as the ending $e$. We extract the sentiment polarity of each sentence by utilizing a lexicon and rule-based sentiment analysis tool (VADER) \cite{conf/icwsm/HuttoG14}:
\begin{equation}
   E_i = \text{VADER}(s_i), i \in [1,5]
\end{equation}
where $E_i$ is a vector of three elements including probabilities of the $i$-th sentence being positive, negative and neutral.

Then, we use a Long Short-Term Memory (LSTM) neural network to encode the sentence sentiments $E_i$  with its context into the hidden state $h_i$, which summarizes the contextual sentiment information around the sentence $s_i$. And we use the last hidden state $h_4$ to predict the sentiment vector $E_p$ in the ending $e$:
\begin{align}
   h_i &= \text{LSTM}(E_i,h_{i-1}) , i\in [1,4] \\
   E_p &= softmax(W_e h_4 + b_e) 
\end{align}

We train the sentiment model by maximizing the cosine similarity between the predicted sentiment vector $E_p$ and the sentiment vector $E_5$ of the correct ending:
\begin{equation}
   sim(S) = \dfrac{E_p \cdot E_5}{\Vert E_p \Vert _2 \cdot \Vert E_5 \Vert _2}
\end{equation} 
Afterwards, we adapt the parameters to the story ending selection task and calculate the following conditional probability $P_S$:
\begin{equation}
    \label{Equ:P_S}
   P_S(y|s_1, ..., s_4, e_i) = softmax(E_pW_sE_e)
\end{equation} 
where $S = \{s_1, s_2, s_3, s_4\}$ is the body, $e_i$ is the candidate ending, $E_p$ is the predicted sentiment vector, $E_e$ is the sentiment vector extracted from ending $e_i$, and $W_s$ is the similarity matrix to be learned.

\subsection{Commonsense Knowledge}
Narrative sequence and sentiment evolution, though useful, are not sufficient to make correct predictions. In a typical story, newly introduced key-words may not be explained in the story because story-writers are not given enough narrative space and time to develop and describe them \cite{narritive}. In fact, there are many hidden relationships among key-words in natural stories. In Figure \ref{Example} (a), although the key-word ``diet" in the ending is not mentioned in the body, there are hidden relationships among ``diet", ``overweight" and ``unhealthy" as shown in Figure \ref{Example} (b).  When this kind of implicit information is uncovered in the model, it is easier to predict the correct story ending.

We leverage the implicit knowledge by using a numberbatch word embedding \cite{speer2017conceptnet}, which is trained on data from ConceptNet, word2vec, GloVe, and OpenSubtitles. The numberbatch achieves good performance on tasks related to commonsense knowledge \cite{DBLP:journals/corr/SpeerL17}. For instance, the cosine similarity between ``diet" and ``overweight" in numberbatch is 0.453, but it is 0.326 in GloVe. This is because numberbatch makes use of the relationship between them as shown in Figure \ref{Example} (b) while GloVe does not.

\begin{algorithm}[h]
  \caption{Knowledge distance computation}
  \begin{algorithmic}[1]
    \ForAll { sentence $s_j$ such that $s_j\in S$}
        \State $distance_j = 0$
        \State $num = 0$
        \ForAll { word $w$ such that $w\in e_i$}
            \State $max_d$ = $0$
            \State $num += 1$
            \ForAll { word $u$ such that $u\in s_j$}
                \If {$stem(w) != stem(u)$}
                    \State $d$ = cosine similarity(w, u)
                    \If {$d > max_d$}
                        $max_d = d$
                    
                    \EndIf
                \EndIf
            \EndFor

            \State $distance_j += max_d$
        \EndFor
        \State $distance_j  /= num$
    \EndFor
\State return $(distance_1, ..., distance_4)$
  \end{algorithmic}
\end{algorithm}

Given the body $S = \{s_1, s_2, s_3, s_4\}$, a candidate ending $e_i$ and the label $y$, we tokenize each sentence using NLTK and Standford's CoreNLP tools \cite{manning-EtAl:2014:P14-5}. After deleting the stop words, we calculate the knowledge distance vector $D$ between the candidate ending and the body by Algorithm 1. We compute the similarity between two key-words using the cosine similarity of their vector space representations in numberbatch. For each sentence $s_i$ in the body, we then quantify the distance with the ending using averaged alignment score of every key-word in the ending. Then we use a linear layer to model the conditional probability $P_C$:
\begin{equation}
    \label{Equ:P_C}
   P_C(y|s_1, ..., s_4, e_i) = softmax(W_dD + b_d)
\end{equation}
where $W_d$ and $b_d$ are parameters in the linear output layer, and $D$ is the four-dimensional distance vector.

\subsection{Combination Gate}
Finally, we predict the story ending by combining the above three sources of information. We utilize the feature vectors $h_M^k$ in the narrative sequence, $E_e$ in the sentiment evolution, and $D$ in the commonsense knowledge and calculate their cosine similarities. Then we concatenate them into a vector $g$. We use a linear layer to model the combination gate and use that gate to combine three conditional probabilities.
\begin{align}
   G &= softmax(W_gg + b_g) \\
   \tilde{P}(y|s_1, ..., s_4, e_i) &= softmax(sum(G \odot [P_N; P_S; P_C]))
\end{align}
where $W_g$ and $b_g$ are parameters in the linear layer, $(P_N, P_S, P_C)$ are the three probabilities modeled in \eqref{Equ:P_N}, \eqref{Equ:P_S} and \eqref{Equ:P_C},  $G$ is the hidden variable that weighs three different conditional probabilities and $\odot$ is element-wise multiplication. 

Finally, since each of the three components ($P_N$, $P_S$ and $P_C$) are either pre-trained on a separate corpus or individually tuned on the task, we fine-tune the entire model in an end-to-end manner by minimizing the following cost:
\begin{equation}
    \tilde{L} = L_{cm}(S) - \lambda * L_{lm}(C) 
\end{equation}
where $L_{cm}(s) = \sum -ylog(\tilde{P})$ is the cross-entropy between the final predicted probability and the true label, $L_{lm}$ is a regularization term of language model cost, and $\lambda$ is the regularization parameter.

\begin{table}[t]
\begin{tabular}{c|c|c}
\hline
\textbf{Sentences} & \textbf{Number of words} & \textbf{Number of key-words}  \\ \hline\hline
$s_1$           & 8.9          & 6.2         \\ \hline
$s_2$           & 9.9           & 6.5          \\ \hline
$s_3$          & 10.2            & 6.7         \\ \hline
$s_4$           & 10.0           & 6.5    \\ \hline
$e_1$           & 10.5            &5.7         \\ \hline
$e_2$           & 10.3            &5.8      \\ \hline
\end{tabular}\caption{The average number of words and key-words exist in ConceptNet in each sentence of the story}\label{Number}
\end{table}

\section{Dataset}
We evaluated our model on ROCStories \cite{W17-0906}, a publicly available collection of commonsense short stories. This corpus consists of 100,000 five-sentence stories. Each story logically follows everyday topics created by Amazon Mechanical Turk (MTurk) workers. These stories contain a variety of commonsense causal and temporal relations between everyday events. Writers also develop an additional 3,742 stories which contain a four-sentence-long body and two candidate endings. The endings were collected by asking MTurk workers to write both a right ending and a wrong ending after eliminating original endings of given short stories. Both endings were required to include at least one character from the main story line and to make logical sense. and were tested on AMT to ensure the quality. The published ROCStories dataset \footnote{\url{http://cs.rochester.edu/nlp/rocstories}} is constructed with ROCStories as a training set that includes 98,162 stories that exclude candidate wrong endings, an evaluation set, and a test set, which have the same structure (1 body + 2 candidate endings) and a size of 1,871. 

We find that the dataset contains 43,095 unique words, and 28,012 key-words in ConceptNet. The average number of words and key-words in ConceptNet for each sentence are shown in Table \ref{Number}. $s_1$, $s_2$, $s_3$ and $s_4$ are four sentences in the body of stories. $e_1$ and $e_2$ are the two candidate endings. A large portion (65\%) of words mentioned in stories are key-words in ConceptNet.  Thus we believe ConceptNet can provide additional information to the model.  

In our experiments, we use a training set which does not have candidate endings to pre-train the sentiment prediction model. For learning to select the right ending, we randomly split 80\% of stories with two candidates endings in ROCStories evaluation set as our training set (1,479 cases), 20\% of stories in ROCStories evaluation set as our validation set (374 cases). And we utilize the ROCStories test set as our testing set (1,871 cases).

\section{Experiments}
\subsection{Baselines}
We use the following models as our baselines:

\textbf{Msap}\cite{schwartz-EtAl:2017:LSDSem}: Msap uses a linear classifier based on language modeling probabilities of the entire story, and utilizes linguistic features of the ending sentences. These ending “style” features include sentence length, word and character n-gram in each candidate ending (independent of story). 


\textbf{HCM}\cite{D17-1168}: HCM uses FC-SemLM \cite{DBLP:conf/acl/PengR16} in order to represent events in the story, learns sentiment trajectories in a form of N-gram language model,  and  uses  topic-words' GloVe to extract topical consistency feature. It uses Expectation-Maximization for training.

\textbf{DSSM}\cite{learning}:
DSSM first uses two deep neural networks to project the context and the candidate endings into the same vector space, and ending choices based on the cosine similarity of the context.

\textbf{Cai}\cite{P17-2097}: Cai uses BiLSTM RNN with attention mechanisms to encode the body and ending of the story separately and uses a cosine similarity between their representations to calculate the score for each ending during selection process.

\textbf{SeqMANN}\cite{C18-1149}: SeqMANN uses a multi-attention neural network and introduces semantic sequence information extracted from FC-SemLM as external knowledge. The embedding layer concatenates five representations including word embedding, character feature, part-of-speech (POS) tagging, sentiment polarity and negation. The model uses DenseNet to match body with an ending.

\textbf{FTLM}\cite{openai}: FTLM solves the stories cloze test by pre-training a language model using a multi-layer transformer on a diverse corpus of unlabeled text, followed by discriminative fine-tuning.

\subsection{Experimental Settings}
We tune the hyper parameters of models on the validation set. Specifically, we set the dimension of LSTM for sentiment prediction to 64. We use a mini-batch size of 8, and Adam to train all parameters. The learning rate is set to 0.001 initially with a decay rate of 0.5 per epoch.

\section{Results}

\begin{table}[t]
\begin{tabular}{cc}
\hline
\textbf{Model}& \textbf{Accuracy(\%)} \\ \hline 

Msap\cite{schwartz-EtAl:2017:LSDSem}          & 75.2               \\ \hline
HCM \cite{D17-1168}          &  77.6       \\ \hline 
DSSM  \cite{learning}        & 58.5                 \\ \hline
Cai \cite{P17-2097}          &74.7   \\ \hline
SeqMANN  \cite{C18-1149}       & 84.7                  \\ \hline
FTLM  \cite{openai}         & 86.5      \\ \hline\hline
Our Model(Plot\&End)     &78.4        \\ \hline
Our Model(Full Story)     &\textbf{87.6*} \\ \hline
\end{tabular}\caption{Performance comparison with baselines, *indicates that the model is significantly better than best baseline model}\label{Result1}
\end{table}

We evaluated baselines and our model using accuracy as the metric on the ROCStories dataset, and summarized these results in Table \ref{Result1}. 
The linear classifier with language model, \textbf{Msap}, achieved an accuracy of 75.2\%. When adding additional features, such as sentiment trajectories and topic words to traditional machine learning methods, \textbf{HCM} achieved an accuracy of 77.6\%. Recently, more neural network-based models are used. \textbf{DSSM} simply used a deep structured semantic model to learn representations for both bodies and endings only achieved an accuracy of 58.5\%.  Utilizing  \textbf{Cai} improved neural model performance to 74.7\% by applying attention mechanisms on a BiLSTM RNN structure. \textbf{SeqMANN} further improved the performance to 84.7\%, when combining more information from embedding layers, like character features, part-of-speech (POS) tagging features, sentiment polarity, negation information and some external knowledge of semantic sequence.
Researchers also improved model performance by pre-training word embeddings on external large corpus. \textbf{FTLM} pre-trained a language model on a large unlabeled corpus and fine-tuned on the ROCStories dataset, and achieved an accuracy of 86.5\%. 

We tried two different ways to construct narrative sequence features: Plot\&End and FullStory. Plot\&End encodes the body and ending of a story separately and then computes their cosine similarity. We use a hierarchy structure to encode the four body sentences. However using such encoding method, our model only achieved an accuracy of 78.4\%. One possible reason is that the relation between sentences learned through pre-trained language models are not fully explored if we encode each sentence separately. FullStory encodes all five sentences together. Our model achieved the best performance when using FullStory mode to encode narrative sequence information. We achieved an accuracy of 87.6\%, outperforming all baseline models. Such improvement may come from the full use of the pre-trained transformer block, as well as the incorporation of the structured commonsense knowledge and sentiment information in the model.

\begin{table}[t]
\centering
\begin{tabular}{cc}
\hline
\textbf{Types of information}  & \textbf{Accuracy(\%)}    \\ \hline
Narrative          & 85.3   \\            
Sentiment           &  58.7   \\     
Knowledge             & 63.8                 \\ \hline
Our Model(All Types)   &\textbf{87.6} \\ \hline

\end{tabular}\caption{Performance on only using one type of information }\label{Result2}
\end{table}

\begin{table}[t]
\centering
\begin{tabular}{cc}
\hline
\textbf{Types of information}  & \textbf{Accuracy(\%)}    \\ \hline
Our Model(All Types)   &\textbf{87.6} \\ \hline
- Narrative          & 65.9               \\ 
- Sentiment           &  87.2       \\  
- Knowledge             & 85.6    \\ \hline

\end{tabular}\caption{Performance on stripping one type of information, e.g. ``- Sentiment" means removing sentiment information.}\label{Result3}
\end{table}

\subsection{Ablation Study}
We conducted another two groups of experiments to investigate the contribution of the three different types of information: narrative sequence, sentiment evolution and commonsense knowledge. First, we measure the accuracy of only using one type of information at a time and describe the result in Table \ref{Result2}. When we use just one type of information, the performances are worse than when using all of the information, suggesting a single type of information is insufficient for story ending selection. We also measure the performance of our model by stripping one type of information at a time and display the results in Table \ref{Result3}. We observe that by removing the narrative sequence information, the model performance decreases most significantly. We suspect this is because the narrative chain is the key element that differentiates a story from other types of writing. Therefore, removing narrative sequence information makes it difficult to predict the story ending. 
If we only use the narrative sequence information, the performance is 85.3\%. When commonsense knowledge is added to the model on top of the narrative sequence information, the performance improves to 87.2\% which is statistically significant. When sentiment evolution information is added, the model only improves to 87.6\%. We speculate this is because the pre-trained language model from narrative sequence information may already capture some sentiment information, as it is trained on an ensemble of several large corpus. 
This suggests that commonsense knowledge has a large impact on narrative prediction task.

\newcommand{\tabincell}[2]{\begin{tabular}{@{}#1@{}}#2\end{tabular}}
\begin{table*}[htb]
\begin{tabular}{|c|c|c|c|}
\hline
  \textbf{Body} & \textbf{Correct ending} & \textbf{Wrong ending}\\
\hline

  \tabincell{c}{Agatha had always wanted pet birds. \\ So one day she purchased two pet finches.\\Soon she couldn’t stand their constant noise. \\ And even worse was their constant mess.}  & \tabincell{c}{Agatha decided to return them. } & \tabincell{c}{Agatha decided to \\ buy two more.} \\
\hline

 \tabincell{c}{Jackson had always wanted to grow a beard. \\ His friends told him that a beard would look bad, but he ignored them.\\Jackson didn't shave for a month and he grew a bushy, thick beard. \\ Admiring himself in the mirror, Jackson felt satisfied.}  & \tabincell{c}{He was glad that he \\ hadn't listened to his friends.} & \tabincell{c}{He was ashamed \\ of himself.} \\
\hline

  \tabincell{c}{I was walking through Central Park on a fall day.  \\  I found a stray dog with a collar. \\I called the number on the collar and talked to the owners.  \\ The owners came to the park to pick up their dog.}  & \tabincell{c}{They thanked me very much \\ for finding their dog.} & \tabincell{c}{They let me keep it.} \\
\hline

\end{tabular} \caption{Three examples from ROCStories. The first column is the body of the story, the second column is the correct ending, and the third column is the wrong ending.}\label{Examples}
\end{table*}

\subsection{Case Study}

We present several examples to describe the decision made at the combination gate. All the examples are shown in Table~\ref{Examples}. 


The first story shows how narrative sequence can be the key in detecting the coherent story ending. This one tells a story of Agatha and birds. As we have analyzed in the narrative sequence, the narrative chain is apparently the most effective clue in deciding the right ending. In the combination gate, the narrative part's weight is 0.5135, which is larger than the sentiment component's weight, 0.2214 as well as the commonsense component's weight of 0.2633. The conditional probability of the correct ending given the narrative information is 0.8634, which is much larger than the wrong ending. As both sentences' sentiments are neutral, the sentiment information is not useful . And as the word ``buy'' has closer relation to ``want" and ``purchase" mentioned in the sentence body than the word,``return", the commonsense knowledge actually makes the wrong decision which gives slightly higher probabilities to the wrong ending(0.5642).

The second story shows why and how sentiment evolution is influencing the final performance. It is a story about Jackson's beard: Jackson wanted to grow a beard regardless of what his friends said, and he was satisfied with his bushy, thick beard. Clearly the emotions between the two candidate endings are different. Based on the rule of consistent sentiment evolution, an appropriate ending should have a positive emotion rather than a negative emotion. The output of our model shows that in the combination gate, the sentiment evolution component received the largest weight, 0.4880, while the narrative sequence and the commonsense knowledge component have a weight of 0.2287 and 0.2833. Finally, the probability of the correct ending is 0.5360, larger than that of the wrong ending which is 0.4640 in sentiment part. Whereas in the narrative sequence component, the probability of the correct option is 0.4640, smaller than the wrong ending which is 0.5360. Other models like FTLM \cite{openai} that only rely on narrative sequence will make the wrong decision in this case. The probabilities of the commonsense knowledge component is 0.5257 versus 0.4725. Through combination gate, our model mainly relies on the sentiment to make a selection. As a result, it will identify the right ending despite  other components influence toward a wrong decision.

\begin{figure}[h]
    
    \includegraphics[width=8cm,height=4cm]{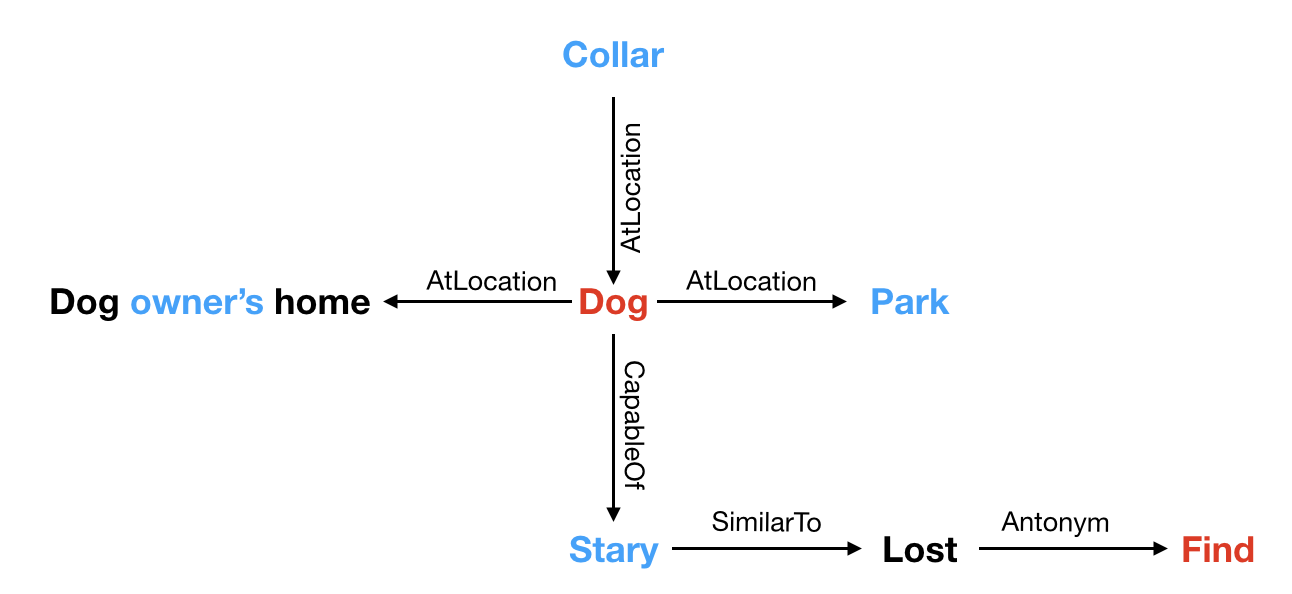}
     \caption{Sub-graph in ConceptNet}\label{KG2}
    
\end{figure}

The third example presents the roles commonsense knowledge plays in our model. It tells a story about a person finding a dog. The sentiments of the two candidates  are both neutral again. But based on the knowledge graph in ConceptNet, shown in Figure \ref{KG2}, there exists many relations between the correct ending and the story body. The key-words in the ending are in red, and the key-words in the story body are in blue.  The key-words such as ``stray" and ``collar" are highly associated with ``dog" and ``find" in the correct ending.  The result shows that the gate gives the commonsense knowledge component a weight of 0.5156, which is the largest among the three components. The conditional probability of the correct ending considering commonsense information (0.5540) is larger than the wrong ending as we expected. In this case, the narrative sequence component makes the wrong decision, which gives higher probabilities to the wrong ending (0.5283). Thus models like FTLM \cite{openai} which only consider narrative chain will identify the wrong ending. However, as the combination gate learns to trust the commonsense knowledge component in this example more, our model still predicts the correct ending.

We can see that our model is able to learn to rely on different information types based on the content of different stories. We obtain such model effectiveness by using a combination gate to fuse all three types of information, and in doing so, understand how all three are imperative in covering all possible variations in the dataset. 

\begin{figure}[t]
    \centering
    \includegraphics[width=8cm,height=3cm]{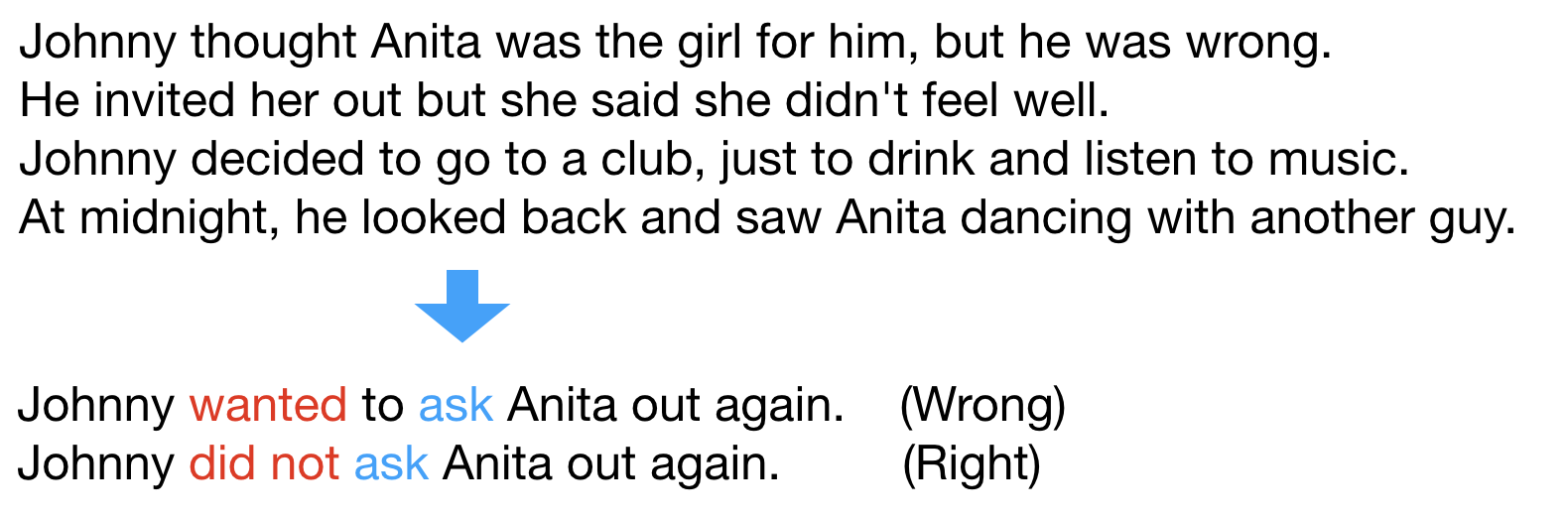}
     \caption{An example involves negation in ROCStories}\label{Negation}
    
\end{figure}



However, it is still challenging for our model to handle the stories that have negations. Figure~\ref{Negation} shows an example. It tells a story between Johnny and Anita. But the only difference between two candidate endings is the negation word. Even when fusing three types of information, our model still cannot get the answer right. Because both event chains are about ``asking Anita out", they are both neutral in sentiment, and the key-words in these two endings are the same as well. In the future, we plan to incorporate natural language inference information to the model to handle such cases.

\section{Conclusion}
Narrative completion is a complex task that requires both explicit and implicit knowledge. We proposed a neural network model that utilized a combination gate to fuse three types of information including: narrative sequence, sentiment evolution and structured commonsense knowledge to predict story endings. The model outperformed state-of-the-art methods. We found that introducing external knowledge such as structured commonsense knowledge helps narrative completion.


\bibliography{mybib}
\bibliographystyle{aaai}
\end{document}